\documentclass[10pt, a4paper]{article}

\usepackage{listings}
\usepackage{seqsplit}
\usepackage{multirow}
\usepackage{amsmath}

\usepackage[final]{lrec2026} 

\title{FHIRPath-QA: Executable Question Answering over FHIR Electronic Health Records}

\name{Michael Frew, Nishit Bheda, Bryan Tripp} 
\address{University of British Columbia; University of Waterloo \\
         michael.frew@ece.ubc.ca, nishit.bheda@uwaterloo.ca, bptripp@uwaterloo.ca\\
         }

\abstract{
Though patients are increasingly granted digital access to their electronic health records (EHRs), existing interfaces may not support precise, trustworthy answers to patient-specific questions. Large language models (LLM) show promise in clinical question answering (QA), but retrieval-based approaches are computationally inefficient, prone to hallucination, and difficult to deploy over real-life EHRs. This work introduces FHIRPath-QA, the first open dataset and benchmark for patient-specific QA that includes open-standard FHIRPath queries over real-world clinical data. A text-to-FHIRPath QA paradigm is proposed that shifts reasoning from free-text generation to FHIRPath query synthesis. For o4-mini, this reduced average token usage by 391$\times$ relative to retrieval-first prompting (629,829 vs 1,609 tokens per question) and lowered failure rates from 0.36 to 0.09 on clinician-phrased questions. Built on MIMIC-IV on FHIR Demo, the dataset pairs over 14k natural language questions in patient and clinician phrasing with validated FHIRPath queries and answers. Empirically, the evaluated LLMs achieve at most 42\% accuracy, highlighting the challenge of the task, but benefit strongly from supervised fine-tuning, with query synthesis accuracy improving from 27\% to 79\% for 4o-mini. These results highlight that text-to-FHIRPath synthesis has the potential to serve as a practical foundation for safe, efficient, and interoperable consumer health applications, and the FHIRPath-QA dataset and benchmark serve as a starting point for future research on the topic. The full dataset and generation code can be accessed on \href{https://github.com/mooshifrew/fhirpath-qa}{GitHub}.
\\ 
\newline 
 }

\begin{document}

\maketitleabstract

\section{Introduction}

Over the past decade, healthcare systems worldwide have adopted standardized, interoperable representations of electronic health records (EHRs). HL7 Fast Healthcare Interoperability Resources (FHIR) has emerged as the dominant framework, with national adoption across major EHR vendors and mandated API requirements in jurisdictions such as the United States \cite{hl7_fhir, world2023and}.

However, digital access does not equal understanding. Although patients can increasingly view their EHRs, FHIR’s graph-structured design makes navigation difficult \cite{zhang2014better, roman2017navigation}. Most systems rely on keyword search, often returning entire documents or dense tables rather than the specific value requested \cite{hanauer2015supporting}. Bridging this gap requires automated question-answering (QA) systems capable of retrieving precise, verifiable answers from a patient’s own record \cite{del2014clinical}.

Large Language Models (LLMs) demonstrate strong clinical reasoning on general medical QA tasks \cite{bardhan2024question}. However, deploying them over patient-specific EHR data introduces substantial challenges. Context windows are finite, making full-record ingestion costly and prone to degradation as length increases \cite{liu2024lost, chen2024benchmarking}. Retrieval-Augmented Generation (RAG) systems remain vulnerable to hallucination \cite{zhang2025hallucination}, which is unacceptable in patient-facing tools where incorrect answers may influence health decisions \cite{bussone2015role}. Additionally, LLMs struggle to reliably traverse FHIR’s deeply nested graph structure; benchmarks such as FHIR-AgentBench report consistent failures in multi-hop reasoning and reference resolution \cite{lee2025fhir}. Finally, transmitting sensitive patient data to external models raises privacy concerns, while hosting sufficiently powerful models locally is often impractical.

To address these limitations, we turn to the standard itself. While FHIR defines the data structure, FHIRPath \cite{hl7_fhirpath} is its native path-based query language. Unlike SQL—which requires backend database access rarely available to patient-facing applications—FHIRPath operates at the API layer and is broadly supported across interoperable health systems.

Adopting a Text-to-FHIRPath approach offers three distinct advantages over RAG or SQL-based methods:
\begin{enumerate}
    \item \textbf{Determinism \& Safety:} A generated query acts as an intermediate logic step that can be validated before execution, substantially reducing hallucination at the answer level by shifting risk to symbolic query generation.
    \item \textbf{Data Minimization:} By executing the query logic locally or at the API level, the system retrieves only the specific data point requested (e.g., a single observation value) rather than retrieving and processing entire clinical documents. This drastically reduces the context window cost and minimizes personally identifiable information exposure.
    \item \textbf{Interoperability:} A model trained to generate FHIRPath works across any system compliant with the US Core Data for Interoperability (USCDI) standard, whereas SQL generation is brittle and tied to specific database schema.
\end{enumerate}

Despite the growing need for patient-centric QA, existing benchmarks largely evaluate adjacent but insufficient problem settings. Much prior work targets clinician-facing analytics or structured querying for secondary clinical use \cite{lee2022ehrsql}, rather than patient access to their own records. Patient-oriented benchmarks such as archEHR-QA emphasize interpreting verbose patient questions, identifying relevant evidence, and generating grounded answers from free-text clinical notes \cite{PhysioNet-archehr-qa-bionlp-task-2025-1.3}. While critical for modeling patient–clinician communication, these tasks operate at the document level and do not evaluate precise, field-level access to structured EHR data. 

Benchmarks that evaluate QA over FHIR often rely on synthetic or heavily curated data \cite{soni2019using, kothari2025question}, substantially reducing the ambiguity, noise, and scale of real patient records. More recent benchmarks assess agentic interaction with live FHIR servers \cite{lee2025fhir}, evaluating end-task correctness but without exposing or supervising the executable query logic required to reliably traverse FHIR graphs.

This work proposes a deterministic FHIRPath-based approach as a safer, verifiable alternative to standard retrieval methods. By pairing executable query logic with answers, the approach prioritizes precision and safety over broad recall. The method is evaluated using a novel benchmark derived from MIMIC-IV on FHIR Demo and EHR-SQL \cite{bennettMIMICIVFHIR,mimicOnFHIRPAPER,lee2022ehrsql, lee2024overview}, leveraging real clinical data rather than synthetic proxies. A patient-perspective paraphrasing pipeline is introduced to assess the system’s ability to translate non-expert language into structured FHIRPath queries. An overview of the primary contributions follows:

\begin{itemize}
    \item \textbf{A Question-FHIRPath-Answer dataset.} FHIRPath-QA combines realistic clinical questions, executable FHIRPath queries,  and answers from real MIMIC-IV on FHIR patient data. The full dataset as well as generation code is publicly available for reproducibility and to support future extension.
\item \textbf{Patient-oriented Paraphrasing.} Real patients do not ask for "tachycardia" status; they ask, "Why is my heart racing?" This work includes a pipeline to translate clinical questions into layperson language, evaluating the model's ability to bridge the semantic gap between vague patient intent and strict FHIR schema.
\item     \textbf{Empirical Evaluation} of retrieval-first and query-first QA pipelines with commercial and fine-tuned models. 
\end{itemize}

\section{Related Work}

\begin{figure}
    \centering
    \includegraphics[width=0.97\linewidth]{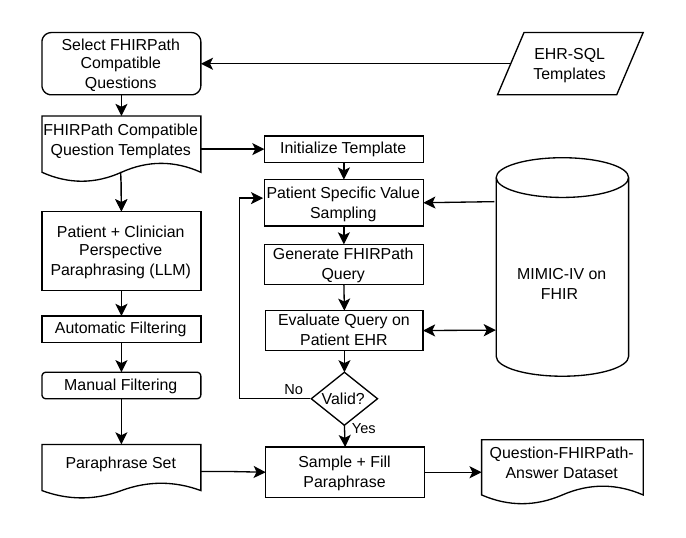}
    \caption{Building FHIRPath-QA involved question selection, LLM-based paraphrasing, and query generation grounded in real patient data.}
    \label{fig:dataflow}
\end{figure}

Early EHR QA focused on Text-to-SQL generation for clinician analytics. MIMICSQL \cite{wang2020text}, EHR-SQL \cite{lee2022ehrsql}, and EHR-SQL-2024 \cite{lee2024overview} established strong baselines using real clinical questions grounded in the MIMIC-III \cite{johnson2016mimic}, eICU \cite{pollard2018eicu} and MIMIC-IV databases \cite{johnson2023mimic}. However, these systems rely on direct database access (SQL), which is typically inaccessible from patient-facing applications. Further, these databases have non-standard schema, so transferability to clinical use is limited.

There are limited benchmarks for QA models on FHIR. \citet{soni2019using} curated one of the first question-query-answer data sets in FHIR data. However, they did so using Synthea \cite{walonoski2018synthea}, a synthetic data-generation pipeline, so the EHRs lack the noise, variability, and scale of real clinical data.  \citet{kothari2025question} similarly used Synthea to generate a QA dataset for fine-tuning LLMs, but these questions were generated after selecting a small set of relevant resources, further simplifying the task.  Consequently, systems are not required to reason over large, longitudinal EHRs with the complexity or scale of real patient records.

More recently, a paradigm shift toward agentic interaction has led to benchmarks that evaluate an LLM’s ability to interact with live FHIR servers. MedAgentBench \cite{jiang2025medagentbench} assesses end-to-end task completion across 10 clinical workflows using 300 physician-authored questions, emphasizing practical correctness. FHIR-AgentBench \cite{lee2025fhir} similarly evaluates agentic performance, but focuses specifically on clinician-oriented question answering. Correctness is determined using annotated keys of relevant resources. While this benchmark meaningfully advances QA evaluation on real FHIR data, it does not expose or supervise the executable query logic used to retrieve answers, as the underlying dataset generation relies on SQL. Consequently, the intermediate reasoning required to traverse FHIR graphs is not directly observable or learnable, highlighting the absence of benchmarks that expose executable query logic over real FHIR data. 

Taken together, prior work either targets clinician-facing analytics via Text-to-SQL, evaluates FHIR-based QA on synthetic or heavily curated data, or measures agentic interaction with FHIR servers without exposing the underlying executable logic. As a result, none directly supervise the translation of natural language questions into verifiable, executable queries over real-world FHIR data. FHIRPath-QA addresses this gap in Text-to-FHIRPath generation by introducing a benchmark that pairs patient and clinician language questions with executable FHIRPath queries over real MIMIC-IV on FHIR records.

\section{Dataset Construction}

\subsection{Data Source}

To ensure questions, queries, and answers are realistic, patient data was used from the MIMIC-IV on FHIR Demo Dataset \cite{bennettMIMICIVFHIR, mimicOnFHIRPAPER}. Though the released dataset draws from the 100-patient EHRs in MIMIC-IV on FHIR Demo, the generation code extends to the full MIMIC-IV dataset. This data was uploaded into a local instance of the HAPI FHIR server as it is a robust open-source implementation of the HL7 FHIR standard \cite{hussain2018learning}.

\subsection{Question Templates}
 
Standardized question templates were adapted from EHR-SQL \cite{lee2022ehrsql} to ensure they have clinical relevancy. Templates were selected that focus on individual patients (e.g. "What was the last prescription for patient X?") rather than population-level or aggregate statistics (e.g. "How many patients were diagnosed with hepatitis this year?"). This choice was made because individual patients would not have access to, nor an interest in, other patients' EHRs. Finally, questions were filtered to ensure they can be answered from the information in FHIR data by a single FHIRPath execution. In total, 61 unique question templates were selected covering a broad range of clinical topics, summarized in \autoref{tab:question_topics}.

\begin{table}[t]
\centering
\begin{tabular}{l p{0.5\linewidth}}
\hline
FHIR Resource & Clinical Topics \\
\hline
Patient & Demographics, birth date, gender \\
Encounter & Hospital visits, ICU stays, admission types, discharge times  \\
Observation & Lab tests, vital signs, weight, intake events, output events, microbiology results \\
Condition & Diagnoses, clinical status \\
Medication (Med.) & Drug names \\
Med.Request & Prescriptions, dosages, drug delivery routes  \\
Med.Administration & Medication administered to patient in hospital \\
Procedure & Clinical interventions, surgical procedures, timing of events \\
\hline
\end{tabular}
\caption{The FHIR resource types used in FHIRPath-QA cover a range of clinical topics required for inpatient care.}
\label{tab:question_topics}
\end{table}

\subsection{Multi-Perspective Paraphrasing}

A multi-perspective paraphrasing strategy was developed to measure the robustness of  models across diverse user groups. While existing datasets like EHR-SQL primarily focus on clinician-oriented language, FHIRPath-QA accounts for the varying degrees of medical literacy among potential users of a FHIR-based QA system. The generated paraphrases are divided into two distinct perspectives:
\begin{enumerate}
    \item \textbf{The Patient/Caregiver Perspective:} Recognizing that patients and caregivers often access their own data through portals, these paraphrases were generated with more conversational and intuitive phrasing. Further, specification of the patient's ID is removed from the question, since it is assumed that this information would be known from user identification. 
    \item \textbf{The Clinician Perspective:} This category maintains the technical precision of the original templates. Formal medical terminology is maintained to reflect the language that would be used by a physician or nurse. 
\end{enumerate}

Following the observation that modern LLMs can outperform traditional language-generation methods, OpenAI gpt-4o was used for paraphrase generation. For each question template, 50 paraphrases were generated from both the clinician and patient perspectives. The refinement process followed a three-step pipeline: 

1) All required variables and fillable placeholders in the query templates were verified to be preserved correctly during paraphrasing.

2) Lexical diversity was enforced by discarding paraphrases with a pairwise Levenshtein distance less than 10 or a normalized distance under 0.15. 

3) Semantic alignment was maintained by encoding sentences with the all-MiniLM-L6-v2 transformer and computing cosine similarity. A lower similarity threshold was applied for patient-perspective paraphrases to accommodate the natural linguistic variance of non-physician language. 

This process removed over half of the total paraphrases, resulting in a final dataset of 1,239 clinician and 856 patient entries. Patient paraphrases exhibited higher lexical variety and lower semantic similarity compared to clinician paraphrases. For details, see Appendix \ref{app:paraphrase-stats}.

\begin{figure}[t]
    \centering
    \includegraphics[width=1.0\linewidth]{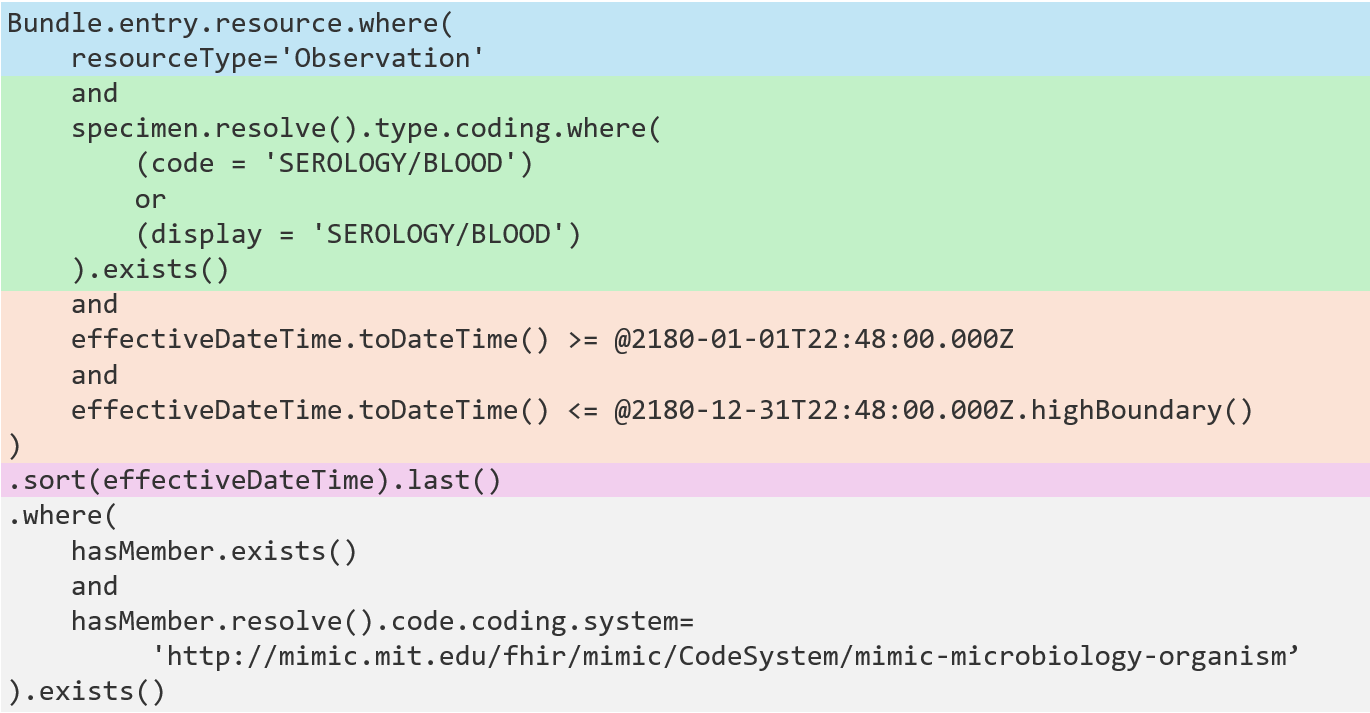}
    \caption{The query generated to answer the question \textit{Were any organisms found in my last SEROLOGY/BLOOD microbiology test this year?} Colored regions indicate logical stages of the query: resource type selection (blue), specimen filtering (green), temporal constraints (orange), and selection of the most recent observation (purple). A final post-aggregation constraint checks that an organism was found (grey).}
    \label{fig:annotated_query}
\end{figure}

\subsection{FHIRPath Query Generation}

Consistent with established datasets, the query template and placeholder structure introduced by \citet{lee2022ehrsql} are adopted. This framework is extended by mapping question templates to the hierarchical FHIR R4 standard, replacing relational SQL joins with path-based FHIRPath traversals.

FHIRPath expressions were crafted to be executed on a patient bundle containing all resources needed to answer the question. This assumption is valid queries are executed on the server side. For local execution, such a scenario can be replicated by running queries on a patient's full bundle: the result of an `\$everything` export. 

The questions can be broadly grouped based on four primary response types: counts (e.g. frequency of interventions), existence (true/false), lists (returning multiple entities), and exact responses (retrieving specific names, dosages, or timestamps). Each question type necessitates a distinct set of navigation steps and filter applications to account for the decentralized nature of the FHIR R4 standard, and the different sets of resources relevant to each question. For instance, querying a microbiology test involves navigating the relationship between the event of specimen collection and the event of the result report. These are recorded as distinct clinical observations and require complex link traversal. An example of this is shown in \autoref{fig:annotated_query}.

\subsection{Query Validation}

The generation pipeline was extensively tested and validated on patient bundles of varying sizes. These tests focused on ensuring syntactic validity of queries by parsing them with an open-sourced, rust-native FHIRPath implementation from OctoFHIR \cite{octofhir_fhirpath_rs}. Queries were manually validated for semantic consistency across multiple placeholder configurations.

Further validation focused on execution accuracy over manually annotated FHIR bundles. Since MIMIC-IV on FHIR patient bundles are very large, five smaller versions (averaging 17 resources per bundle) were curated for this task to ensure each question could be answered. Each query type was validated over the five curated patient bundles.

\subsection{Dataset Assembly}

To ensure questions have answers (e.g. not asking about ICU visits if the patient has never been in the ICU), questions and queries were randomly generated based on the context of specific patients' entire EHR. To account for date shifts in the de-identification process of MIMIC-IV, datetime filtering was computed with the current date being the date of the last record in the EHR. Queries were then executed against patient bundles to generate the corresponding answers.

\begin{table*}[t]
\centering
\begin{tabular}{lcccc}
\hline
\multirow{2}{*}{\textbf{Split}} &\multicolumn{2}{c}{\textbf{Benchmark}} & \multicolumn{2}{c}{\textbf{Large}} \\
\cline{2-5}
& \textit{Clinician} & \textit{Patient} & \textit{Clinician} & \textit{Patient} \\
\hline
\textbf{Train}  & 954 (45.5\%) & 675 (32.2\%) & 4721 (38.7\%) & 4775 (39.1\%) \\
\textbf{Val}  & 104 (5.0\%) & 89 (4.2\%) & 535 (4.4\%) & 661 (5.4\%) \\
\textbf{Test}  & 181 (8.6\%) & 92 (4.4\%) & 791 (6.5\%) & 717 (5.9\%) \\
\hline
\textbf{Total}  & \textbf{1239 (59.1\%)} & \textbf{856 (40.9\%)} & \textbf{6047 (49.6\%)} & \textbf{6153 (50.4\%)} \\
\hline
\textbf{Combined} & \multicolumn{2}{c}{\textbf{2095}} & \multicolumn{2}{c}{\textbf{12200}} \\
\hline
\end{tabular}
\caption{Dataset Statistics. There are 100 distinct patient records and 61 question templates. The Benchmark contains one question, query, and executed answer per unique paraphrase. FHIRPath-QA Large contains multiple questions and queries per paraphrase, without executed answers. Samples are stratified by split and further categorized by perspective (Clinician vs. Patient). The bottom row displays the combined counts across both perspectives.}
\label{tab:dataset_stats}
\end{table*}

\subsection{Evaluation Splits}

To evaluate different forms of generalization, three complementary evaluation settings were defined.

The dataset is partitioned into training (78\%), validation (9\%), and test (13\%) splits using a stratified split over natural language paraphrases. Paraphrases are disjoint across splits, while question templates and patients may overlap. This setting evaluates linguistic generalization and is used for all baseline comparisons.

A \textit{query} holdout is defined, in which four entire question templates are excluded from training. These templates correspond to executable FHIRPath queries that do not appear elsewhere in the dataset, while still operating over FHIR resource types observed during training. It accounts for 155 samples in FHIRPath-QA Benchmark. Performance on this split evaluates a model’s ability to synthesize novel executable query logic from natural language, rather than relying on memorized template-level mappings.

To assess structural generalization beyond the observed schema, a \textit{resource} holdout is defined consisting of five templates that require FHIR resources not present in the training data, specifically \texttt{Location} and \texttt{MedicationAdministration}. This holdout is composed of 126 samples in FHIRPath-QA Benchmark. Success on this split indicates whether fine-tuning enables generalization of FHIRPath syntax and traversal patterns to unseen resource types, despite no exposure to their schemas during training.

\subsection{Dataset Makeup}

\noindent \textbf{Resources:} The FHIRPath-QA dataset is released as a two-tier resource designed to support both model training and reliable execution-based evaluation:

\begin{itemize}
    \item \textbf{FHIRPath-QA Benchmark:} A high-precision evaluation set containing 2,095 question-query-answer triples. Every entry in this subset has been validated via execution against the patient's `\$everything` bundle to ensure correctness. This serves as an execution-validated benchmark for measuring execution accuracy and linguistic generalization.
    
    \item \textbf{FHIRPath-QA Large:} A large-scale corpus containing 12,200 question-query pairs. The queries in this dataset are grounded in real patient records, but answers are not provided via execution. This resource is designed specifically for supervised fine-tuning (SFT), providing the high volume of examples needed for models to learn the complex syntactic nesting of FHIRPath.
\end{itemize}

\noindent \textbf{Volume and Diversity:} In total, the dataset comprises 14,295 unique samples derived from 61 question templates and 2,095 paraphrases (\textasciitilde34.3 per template) which are grounded in 100 real patient EHRs. Each sample is annotated with the paraphrase split label and an indicator for belonging in the unseen query and resource-level holdout groups for simple filtering that ensures large-scale training sets are disjoint from the evaluation benchmarks.

\section{Experiments}

Experiments were designed and carried out with the following goals in mind: (1) to compare the performance and token-efficiency trade-offs between query-first (text-to-FHIRPath) and retrieval-first approaches; (2) to assess the robustness of these approaches to the variations in patient and clinician-speech; and (3) to investigate the extent to which SFT enables generalization across novel paraphrases, unseen FHIRPath logic, and novel FHIR resource types. All evaluation results in this paper are reported on FHIRPath-QA Benchmark.

\subsection{Architectures}
Two distinct architectural paradigms for clinical question answering were evaluated. 
\begin{enumerate}
    \item \textbf{Ingestion-based Retrieval} approach (referred to as "retrieval") modeled after the single-turn agent in FHIRAgentBench \cite{lee2025fhir}, requires the LLM to identify and retrieve relevant FHIR resources for a given patient and perform reasoning over the raw content. 
    \item \textbf{FHIRPath query-based execution} approach (referred to as "query-first") tasks the agent with generating a FHIRPath query. This query is then executed against the patient’s complete FHIR bundle to extract an answer, shifting the computational burden from LLM inference over long contexts to deterministic query execution.
\end{enumerate}

\subsection{Metrics}

Final answer accuracy was used to evaluate model performance. This is assessed as an exact match of the executed result for the query-first approach. Since the retrieval-based approach produces free-text answers, responses were manually evaluated for correctness against the execution-validated benchmark by two independent annotators (the first and second author), both with graduate-level training and relevant domain experience. Any disagreements were resolved through discussion and consensus.

Because all experiments were conducted in the single-turn setting, failures occurred for both approaches. For large patient EHRs, the retrieval-first approach occasionally exceeded the model context limit.  For the query-first approach, failures corresponded to invalid FHIRPath syntax. The accuracy excluding these failures was also measured. When failures were included, they were treated as an incorrect answer. 

Efficiency was measured via token usage including prompt and completion tokens. For failed retrieval-based attempts, where the token limit was exceeded, the attempted input tokens was recorded. 

\subsection{Standard Model Comparison}

Baseline performance and efficiency were first evaluated across four foundational models: OpenAI o4-mini, 4.1-mini, 4o-mini, and 4.1-nano. The o4-mini model was included due to strong reported performance in similar evaluations \cite{lee2025fhir}, 4.1-mini for its extended context length, 4o-mini as a broadly capable baseline for comparison, and 4.1-nano for its smaller size. Each model was tested using both retrieval-first and query-first architectures. Evaluation was conducted on the test split of FHIRPath-QA Benchmark, as detailed in \autoref{tab:dataset_stats}, establishing performance benchmarks for standard large language models (LLMs) on clinical FHIR tasks.

\subsection{Fine-tuning Protocol} 

To evaluate the impact of specialized training, 4o-mini, 4.1-mini, and 4.1-nano models were fine-tuned on the training samples of FHIRPath-QA Large excluding all samples associated with the unseen query structure and resource-level holdout templates. Training was conducted through the OpenAI API (1 epoch, batch size 5, learning rate 1.8 for 4o-mini, 2.0 for 4.1-mini, and 0.1 for 4.1-nano). Evaluation was conducted on paraphrases in the test split to assess linguistic generalization (Test 1), on the unseen queries to assess an improved understanding of FHIR elements (Test 2), and finally on unseen resource types to assess resource-level generalization (Test 3). 

To explore the effect of the number of patients represented by the training examples, SFT was performed with three subsets of the training dataset containing: 1) 10 patients, 2) 30 patients, and 3) the full 100 patients in MIMIC-IV on FHIR Demo. 

\section{Results} 

\subsection{FHIRPath-QA Stresses Scalability}

\autoref{tab:token_efficiency} shows the token usage per question for reasoning and non-reasoning LLMs for the retrieval-first and query-first models.  Using o4-mini, the query-first approach yielded a 391x reduction in average tokens per question compared to the retrieval-based approach. The retrieval strategy also exhibited extreme variance in token usage ($\text{SD} \approx 1.9 \text{M}$ tokens), making its running-costs less predictable for clinical deployment. This variance scaled directly with EHR size. In contrast, query-first token usage was stable ($\text{SD} \approx 420$ tokens), as the LLM only processes the natural language question and a system prompt. Non-reasoning models had lower token usage, but this had a small impact in the retrieval-first setting, where input tokens dominated.

\begin{table}[h]
\centering
\begin{tabular}{llrr}
\hline
\textbf{Strategy} & \textbf{Model} & \textbf{Tokens} & \textbf{SD} \\ \hline
Retrieval & o4-mini & 629,839 & 1,926,723 \\
Retrieval & 4.1-mini & 577,281 & 1,889,508 \\
\hline
Query-first & o4-mini& 1,609 & 420 \\
Query-first & 4.1-mini& 697 & 24 \\
\hline
\end{tabular}
\caption{The retrieval-first approach consumed a substantially higher number of tokens per question than the query-first approach.}
\label{tab:token_efficiency}
\end{table}

\begin{table*}[t]
\centering
\begin{tabular}{l @{\extracolsep{12pt}} cc cc cc}
\hline
\textbf{Model} & \multicolumn{2}{c}{\textbf{Accuracy}} & \multicolumn{2}{c}{\textbf{Failure Rate}} & \multicolumn{2}{c}{\textbf{Acc. Excl. Failures}} \\ [0ex] \cline{2-7}
\rule{0pt}{1ex} & \textit{Clinical} & \textit{Patient} & \textit{Clinical} & \textit{Patient} & \textit{Clinical} & \textit{Patient} \\ [0ex] 
\rule{0pt}{0ex}\textit{a) Retrieval} & & &  & & & \\ 
\hline
o4-mini & 0.33 & 0.33 & 0.36 & 0.34 & 0.51 & 0.49 \\
4.1-mini  & 0.42 & 0.41 & 0.17 & 0.14 & 0.50 & 0.48 \\
4o-mini & 0.24 & 0.15 & 0.46 & 0.40 & 0.45 & 0.25 \\
\rule{0pt}{0ex}\textit{b) Query-first} & & &  & & & \\ 
\hline
o4-mini & 0.35 & 0.38 & 0.09 & 0.15 & 0.39 & 0.45 \\ 
4.1-mini  & 0.25 & 0.27 & 0.15 & 0.23 & 0.29 & 0.35 \\
4o-mini & 0.25 & 0.28 & 0.02 & 0.09 & 0.26 & 0.31 \\
\hline
\end{tabular}
\caption{Overall accuracy slightly favors the retrieval-first approach, but this metric is influenced by its higher failure rate. When excluding failures, the retrieval-first approach consistently outperforms the query-first approach. Retrieval-first performs better on clinically phrased questions, whereas the query-first approach performs better on patient-style paraphrases.}
\label{tab:performance_comparison}
\end{table*}

\subsection{A Challenging Benchmark}
\label{sec:results-clinician-vs-patient}

\autoref{tab:performance_comparison} lists the results of the retrieval-first and query-first approaches with standard LLMs. Overall, the results show that the FHIRPath-QA benchmark poses a challenge to standard LLMs and agentic architectures, with none achieving an overall accuracy above 42\%. Across models,  o4-mini outperformed 4.1-mini and 4o-mini following the query-first and retrieval-first approaches. When relevant EHRs successfully fit within the context window (i.e. excluding failures), the retrieval-first strategy demonstrated superior baseline performance, which suggests that foundational LLMs lack a thorough understanding of FHIR structures and as a result have a poor ability to convert questions to correct FHIRPath queries.  

Notable differences in performance based on question phrasing and system architecture were observed. Retrieval-first scored more highly on clinician over patient paraphrases, which suggests that foundation models rely heavily on lexical overlap. Technical terms in clinician queries may act as anchors to the formal strings found within the FHIR resource JSONs, allowing for more accurate "needle-in-a-haystack" retrieval. Conversely, in the query-first setting, patient-perspective questions yielded higher accuracy.

\subsection{Impact of Supervised Fine-Tuning}

\begin{figure*}[t]
    \centering
    \includegraphics[width=0.95\linewidth]{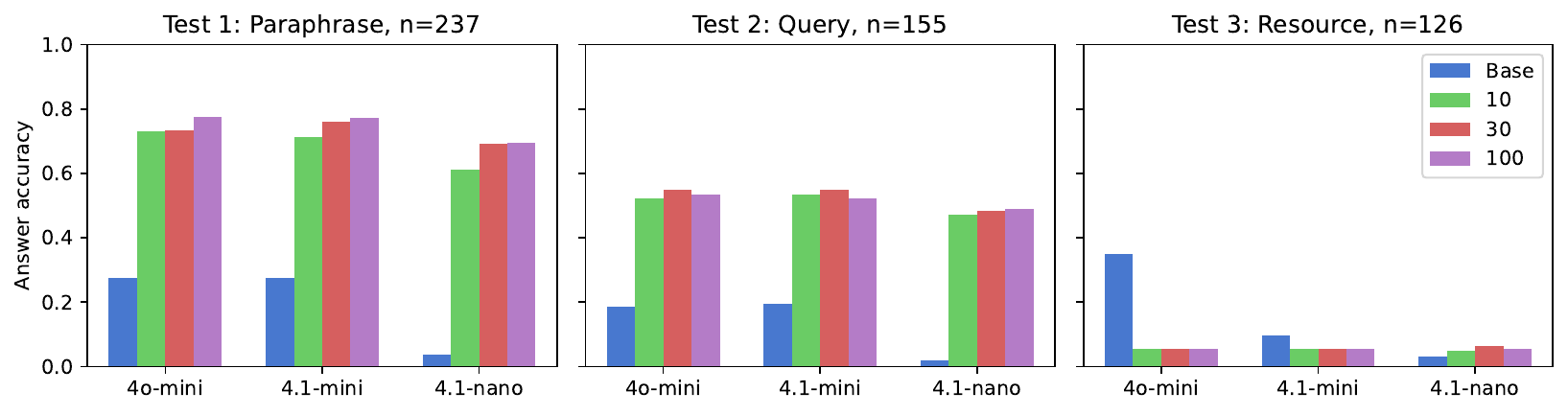}
    \caption{SFT for the query-first approach improved executed answer accuracy on questions with paraphrased inputs over shared query structures (Test 1: Paraphrase) and on questions requiring novel query compositions over FHIR resource types seen during fine-tuning (Test 2: Query). In contrast, performance did not generalize to questions involving unseen resource types (Test 3: Resource). Increasing the number of patients in the fine-tuning dataset (10, 30, 100) led to progressively larger performance gains.}
    \label{fig:FT-combined}
\end{figure*}

Results from SFT on the FHIRPath-QA Large corpus indicate that the apparent performance ceiling of the query-first approach stems from base model pretraining rather than architectural limitations. As shown in \autoref{fig:FT-combined}, SFT leads to a dramatic improvement in the model's ability to map natural language to executable logic. For Test 1, which evaluated novel paraphrases of questions seen in training, all models experienced strong gains. Notably, 4o-mini accuracy nearly tripled, rising from a baseline 27\% to 79\%. This result confirms that specialized training substantially improves the LLMs’ ability to recognize diverse linguistic patterns and map them to executable FHIRPath logic.

Test 2 demonstrates model capacity for logical generalization, which utilized entirely novel question templates composed of clinical reasoning elements seen in other contexts. Performance gains across all models in this category suggest that SFT enabled the agents to learn compositional mappings between clinical intent and the FHIR schema, rather than simply memorizing question–query pairs. 

The results of Test 3 highlight a limit of text-to-FHIRPath generalization. As the models were tasked with querying resources not present in the training set, performance decreased following fine-tuning. While the models become highly proficient in the specific resource types and query patterns represented in the training data, they may overfit to that subset and lose some of the broader, general-purpose schema knowledge present in the base models. 

The scale of patient-specific data used during training also proved to be a critical factor in model proficiency. By comparing models fine-tuned on subsets of 10, 30, and 100 patient records, a consistent upward trend in execution accuracy as the diversity of patient data increased was observed. This suggests that the current performance has not yet reached a plateau; rather, it indicates that future work extending the fine-tuning dataset to the full MIMIC-IV-FHIR corpus would likely yield further gains in handling the natural variance and "noise" found in real-world longitudinal records.

Finally, a notable shift occurred in the model's sensitivity to paraphrase perspective post SFT. While the base models performed better on patient paraphrases from their structural simplicity, fine-tuned models consistently reversed this trend, achieving higher accuracy on clinician-paraphrases (see Appendix \ref{app:tab-combined}). This implies that the inherent ambiguity of layperson language characterized by lower semantic similarity to the clinical ground truth imposes a natural performance ceiling on single-turn systems.

\subsection{Error Analysis}

To evaluate the underlying causes of performance degradation in the query generation pipeline, we performed a qualitative error analysis. Examples of each failure mode are provided in \autoref{app:examples}. 

\subsubsection{Resource Profile Complexities}
A common error mode identified in generated queries stems from an apparent misunderstanding of various FHIR resource profiles. Agents frequently attempt to reference resource elements that do not exist. This failure is indicative that the structure of the resource is not fully understood by the model. For instance, Location resources do not contain temporal information, but some queries tried to access it in search of the time period of a specific visit (Appendix \ref{ssub:example_2}). The correct way to access these types of resources would be to resolve the reference to it, typically via the Encounter resource.

\subsubsection{Semantic Misinterpretation}
A common failure mode arose when the models were faced with question ambiguity, resulting in syntactically correct FHIRPath that does not correctly address the question. This challenge was more prominent in patient-paraphrased questions lacking precise clinical language. For example, in questions inquiring about a patient's medications, a particular challenge is in differentiating MedicationRequest resources (prescriptions) and MedicationAdministration (in-hospital drug administration) resources (see Appendix \ref{ssub:example_3}). Resolving this ambiguity often requires clinical knowledge of whether certain drugs are typically administered in hospital or prescribed. Other subtle misinterpretations include counting total events instead of unique events (Appendix \ref{ssub:example_4}) or considering in-hospital transfers as unique visits.

\subsubsection{Date Filter Mismatch} Up to 30\% of errors in some fine-tuned models were characterized by the generation of the correct FHIRPath query, with the exception of the date filter. In these instances, the model successfully identifies the required resource type and filtering criteria but incorrectly computes the date relative to the current time. An example is provided in Appendix \ref{ssub:example_1}. Quantification of the frequency of this error for each model is provided in Appendix \ref{tab:exact-query-match-accuracy}.

\section{Conclusion}

This work presents FHIRPath-QA, a novel dataset and benchmark combining patient-specific questions, validated FHIRPath queries, and executed answers. It is derived from the MIMIC-IV on FHIR Demo dataset, with clinically relevant questions expressed from both clinician and patient perspectives. To our knowledge, this is the first publicly available benchmark and dataset for Text-to-FHIRPath generation grounded in real EHRs.

Empirical comparison of retrieval-first and query-first QA pipelines shows that query-first approaches offer practical advantages in reliability and efficiency when operating over large EHRs. Executing symbolic queries avoids context window limitations, reduces token consumption by several orders of magnitude, and provides a clear and interpretable failure mode when query generation is unsuccessful. At the same time, empirical results demonstrate that retrieval-based approaches remain competitive in answer accuracy when relevant resources can be ingested, underscoring that query-first execution is not a universal replacement for retrieval-based QA.

These findings show that mapping natural language questions to executable FHIRPath logic remains challenging for pretrained LLMs. However, SFT on question–query pairs substantially improves performance and enables generalization to novel paraphrases and previously unseen question compositions over known FHIR resource types.

Importantly, these results reveal a fundamental tension in patient-facing question answering. While fine-tuning improves executable query synthesis, patient-perspective questions — characterized by greater lexical diversity and semantic ambiguity — impose a natural performance ceiling. Even when models learn FHIRPath syntax, resolving layperson intent into precise clinical constructs remains difficult. This suggests that patient-facing QA may require mechanisms beyond single-turn query generation, such as clarification strategies or schema-aware grounding, to bridge the semantic gap between informal language and structured EHR representations.

By exposing executable query logic over real FHIR data and highlighting the linguistic challenges inherent in patient language, FHIRPath-QA establishes a foundation for safer, more transparent, and more interoperable patient-facing clinical QA systems.

\section{Acknowledgements}

This work was supported by Smile Digital Health. 

\section{Limitations}

This dataset is constructed from MIMIC-IV on FHIR Demo and primarily reflects ICU and in-patient care from a single institution, which may limit generalizability to other clinical settings. Although patient-oriented paraphrases are included, questions are synthetically generated from templates and may not fully capture the variability and scope of real patient-authored queries. The scope of FHIRPath-QA is restricted to factual, record-grounded questions that can be answered deterministically from structured FHIR resources; interpretive or explanatory clinical questions are out of scope. Finally, SFT experiments cover a limited set of resource types, and reduced performance on unseen resources suggests potential over-specialization.

\section{Bibliographical References}\label{sec:reference}

\bibliographystyle{lrec2026-natbib}
\bibliography{references}

@article{bennettMIMICIVFHIR,
  author = {Bennett, Alex and Wiedekopf, Joshua and Ulrich, Hannes and {van Damme}, Philip and Szul, Piotr and Grimes, John and Johnson, Alistair},
  title = {{MIMIC-IV on FHIR}},
  journal = {{PhysioNet}},
  year = {2024},
  month = nov,
  note = {Version 2.1},
  doi = {10.13026/rrj1-ny66},
  url = {https://doi.org/10.13026/rrj1-ny66}
}

@article{mimicOnFHIRPAPER,
    author = {Bennett, Alex M and Ulrich, Hannes and van Damme, Philip and Wiedekopf, Joshua and Johnson, Alistair E W},
    title = {{MIMIC-IV} on {FHIR}: converting a decade of in-patient data into an exchangeable, interoperable format},
    journal = {Journal of the American Medical Informatics Association},
    volume = {30},
    number = {4},
    pages = {718-725},
    year = {2023},
    month = {01},
    issn = {1527-974X},
    doi = {10.1093/jamia/ocad002},
    url = {https://doi.org/10.1093/jamia/ocad002},
    eprint = {https://academic.oup.com/jamia/article-pdf/30/4/718/49524369/ocad002.pdf},
}

@inproceedings{lee2022ehrsql,
  title={{EHRSQL}: A practical text-to-{SQL} benchmark for electronic health records},
  author={Lee, Gyubok and Hwang, Hyeonji and Bae, Seongsu and Kwon, Yeonsu and Shin, Woncheol and Yang, Seongjun and Seo, Minjoon and Kim, Jong-Yeup and Choi, Edward},
  series={NeurIPS 2022}, 
  url={http://dx.doi.org/10.52202/068431-1134}, DOI={10.52202/068431-1134}, 
  booktitle={Advances in Neural Information Processing Systems 35}, 
  publisher={Neural Information Processing Systems Foundation, Inc. (NeurIPS)}, 
  year={2022}, pages={15589–15601}, collection={NeurIPS 2022} 
}

@article{hussain2018learning,
  title={Learning HL7 {FHIR} using the {HAPI} {FHIR} server and its use in medical imaging with the {SIIM} dataset},
  volume={31}, ISSN={1618-727X}, url={http://dx.doi.org/10.1007/s10278-018-0090-y}, DOI={10.1007/s10278-018-0090-y}, number={3}, journal={Journal of Digital Imaging}, publisher={Springer Science and Business Media LLC}, author={Hussain, Mohannad A. and Langer, Steve G. and Kohli, Marc}, year={2018}, month=may, pages={334–340} 
}

@inproceedings{lee2024overview,
  title={Overview of the EHRSQL 2024 Shared Task on Reliable Text-to-SQL Modeling on Electronic Health Records}, url={http://dx.doi.org/10.18653/v1/2024.clinicalnlp-1.62}, DOI={10.18653/v1/2024.clinicalnlp-1.62}, booktitle={Proceedings of the 6th Clinical Natural Language Processing Workshop}, publisher={Association for Computational Linguistics}, author={Lee, Gyubok and Kweon, Sunjun and Bae, Seongsu and Choi, Edward}, year={2024}, pages={644–654} 
}

@article{kothari2025question,
  title={Question Answering on Patient Medical Records with Private Fine-Tuned {LLMs}},
  author={Kothari, Sara and Gupta, Ayush},
  journal={arXiv preprint arXiv:2501.13687},
  year={2025},
  
}

@inproceedings{lee2025fhir,
  title={{FHIR-AgentBench}: Benchmarking {LLM} Agents for Realistic Interoperable {EHR} Question Answering},
  author={Lee, Gyubok and Bach, Elea and Yang, Eric and Pollard, Tom J. and Johnson, Alistair and Choi, Edward and Jia, Yugang and Lee, Jong Ha},
  booktitle={Proceedings of the Machine Learning for Healthcare Conference (ML4H)},
  year={2025},
  url={https://openreview.net/forum?id=Le1hGVQNb8}
}

@article{jiang2025medagentbench,
  title={MedAgentBench: A Virtual EHR Environment to Benchmark Medical LLM Agents}, volume={2}, ISSN={2836-9386}, url={http://dx.doi.org/10.1056/aidbp2500144}, DOI={10.1056/aidbp2500144}, number={9}, journal={NEJM AI}, publisher={Massachusetts Medical Society}, author={Jiang, Yixing and Black, Kameron C. and Geng, Gloria and Park, Danny and Zou, James and Ng, Andrew Y. and Chen, Jonathan H.}, year={2025}, month=aug 
}

@inproceedings{soni2019using,
  title={Using {FHIR} to construct a corpus of clinical questions annotated with logical forms and answers},
  author={Soni, Sarvesh and Gudala, Meghana and Wang, Daisy Zhe and Roberts, Kirk},
  booktitle={AMIA Annual Symposium Proceedings},
  pages={1207--1215},
  year={2019},
  organization={American Medical Informatics Association},
  url={https://pmc.ncbi.nlm.nih.gov/articles/PMC7153115/}
}

@article{bardhan2024question,
  title={Question Answering for Electronic Health Records: Scoping Review of Datasets and Models}, volume={26}, ISSN={1438-8871}, url={http://dx.doi.org/10.2196/53636}, DOI={10.2196/53636}, journal={Journal of Medical Internet Research}, publisher={JMIR Publications Inc.}, author={Bardhan, Jayetri and Roberts, Kirk and Wang, Daisy Zhe}, year={2024}, month=oct, pages={e53636}
}

@misc{hl7_fhir,
  author       = {{HL7 International}},
  title        = {{FHIR}: Fast Healthcare Interoperability Resources},
  howpublished = {\url{https://www.hl7.org/fhir/}},
  note         = {Accessed: 2026-01-22}
}

@misc{hl7_fhirpath,
  author       = {{HL7 International}},
  title        = {HL7 Cross-Paradigm Specification: {FHIRPath}, Release 1},
  howpublished = {\url{https://hl7.org/fhirpath/N1/}},
  year         = {2020},
  note         = {Accessed: 2026-01-22}
}

@book{zhang2014better,
  title={Better {EHR}: usability, workflow and cognitive support in electronic health records},
  author={Zhang, Jiajie and Walji, Muhammad},
  year={2014},
  isbn={978-0-692-26296-2},
  doi={10.13140/2.1.1921.1841},
  publisher={National Center for Cognitive Informatics \& Decision Making in Healthcare}
}

@article{roman2017navigation,
  title={Navigation in the electronic health record: A review of the safety and usability literature}, volume={67}, ISSN={1532-0464}, url={http://dx.doi.org/10.1016/j.jbi.2017.01.005}, DOI={10.1016/j.jbi.2017.01.005}, journal={Journal of Biomedical Informatics}, publisher={Elsevier BV}, author={Roman, Lisette C. and Ancker, Jessica S. and Johnson, Stephen B. and Senathirajah, Yalini}, year={2017}, month=mar, pages={69–79}
}

@article{hanauer2015supporting,
  title={Supporting information retrieval from electronic health records: A report of University of Michigan’s nine-year experience in developing and using the Electronic Medical Record Search Engine ({EMERSE})}, volume={55}, ISSN={1532-0464}, url={http://dx.doi.org/10.1016/j.jbi.2015.05.003}, DOI={10.1016/j.jbi.2015.05.003}, journal={Journal of Biomedical Informatics}, publisher={Elsevier BV}, author={Hanauer, David A. and Mei, Qiaozhu and Law, James and Khanna, Ritu and Zheng, Kai}, year={2015}, month=jun, pages={290–300} 
}

@article{del2014clinical,
  title={Clinical Questions Raised by Clinicians at the Point of Care: A Systematic Review}, volume={174}, ISSN={2168-6106}, url={http://dx.doi.org/10.1001/jamainternmed.2014.368}, DOI={10.1001/jamainternmed.2014.368}, number={5}, journal={{JAMA} Internal Medicine}, publisher={American Medical Association ({AMA})}, author={Del Fiol, Guilherme and Workman, T. Elizabeth and Gorman, Paul N.}, year={2014}, month=may, pages={710}
}

@inproceedings{bussone2015role,
  title={The Role of Explanations on Trust and Reliance in Clinical Decision Support Systems}, url={http://dx.doi.org/10.1109/ichi.2015.26}, DOI={10.1109/ichi.2015.26}, booktitle={2015 International Conference on Healthcare Informatics}, publisher={IEEE}, author={Bussone, Adrian and Stumpf, Simone and O’Sullivan, Dympna}, year={2015}, month=oct, pages={160–169}
}

@article{walonoski2018synthea,
  title={Synthea: An approach, method, and software mechanism for generating synthetic patients and the synthetic electronic health care record}, volume={25}, ISSN={1527-974X}, url={http://dx.doi.org/10.1093/jamia/ocx079}, DOI={10.1093/jamia/ocx079}, number={3}, journal={Journal of the American Medical Informatics Association}, publisher={Oxford University Press (OUP)}, author={Walonoski, Jason and Kramer, Mark and Nichols, Joseph and Quina, Andre and Moesel, Chris and Hall, Dylan and Duffett, Carlton and Dube, Kudakwashe and Gallagher, Thomas and McLachlan, Scott}, year={2017}, month=aug, pages={230–238}
}

@inproceedings{wang2020text,
 series={WWW ’20}, title={Text-to-SQL Generation for Question Answering on Electronic Medical Records}, url={http://dx.doi.org/10.1145/3366423.3380120}, DOI={10.1145/3366423.3380120}, booktitle={Proceedings of The Web Conference 2020}, publisher={ACM}, author={Wang, Ping and Shi, Tian and Reddy, Chandan K.}, year={2020}, month=apr, pages={350–361}, collection={WWW ’20} 
}

@article{johnson2016mimic,
title={{MIMIC-III}, a freely accessible critical care database}, volume={3}, ISSN={2052-4463}, url={http://dx.doi.org/10.1038/sdata.2016.35}, DOI={10.1038/sdata.2016.35}, number={1}, journal={Scientific Data}, publisher={Springer Science and Business Media LLC}, author={Johnson, Alistair E.W. and Pollard, Tom J. and Shen, Lu and Lehman, Li-wei H. and Feng, Mengling and Ghassemi, Mohammad and Moody, Benjamin and Szolovits, Peter and Anthony Celi, Leo and Mark, Roger G.}, year={2016}, month=may
}

@article{johnson2023mimic,
  title={MIMIC-IV, a freely accessible electronic health record dataset}, volume={10}, ISSN={2052-4463}, url={http://dx.doi.org/10.1038/s41597-022-01899-x}, DOI={10.1038/s41597-022-01899-x}, number={1}, journal={Scientific Data}, publisher={Springer Science and Business Media LLC}, author={Johnson, Alistair E. W. and Bulgarelli, Lucas and Shen, Lu and Gayles, Alvin and Shammout, Ayad and Horng, Steven and Pollard, Tom J. and Hao, Sicheng and Moody, Benjamin and Gow, Brian and Lehman, Li-wei H. and Celi, Leo A. and Mark, Roger G.}, year={2023}, month=jan
}

@article{pollard2018eicu,
  title={The {eICU} Collaborative Research Database, a freely available multi-center database for critical care research},
  volume={5}, ISSN={2052-4463}, url={http://dx.doi.org/10.1038/sdata.2018.178}, DOI={10.1038/sdata.2018.178}, number={1}, journal={Scientific Data}, publisher={Springer Science and Business Media LLC}, author={Pollard, Tom J. and Johnson, Alistair E. W. and Raffa, Jesse D. and Celi, Leo A. and Mark, Roger G. and Badawi, Omar}, year={2018}, month=sep 
}

@article{liu2024lost,
  title={Lost in the Middle: How Language Models Use Long Contexts}, volume={12}, ISSN={2307-387X}, url={http://dx.doi.org/10.1162/tacl_a_00638}, DOI={10.1162/tacl_a_00638}, journal={Transactions of the Association for Computational Linguistics}, publisher={MIT Press}, author={Liu, Nelson F. and Lin, Kevin and Hewitt, John and Paranjape, Ashwin and Bevilacqua, Michele and Petroni, Fabio and Liang, Percy}, year={2024}, pages={157–173}
}

@article{world2023and,
  title={{WHO} and {HL7} collaborate to support adoption of open interoperability standards},
  author={World Health Organization and others},
  journal={World Health Organization. Recuperado el},
  volume={10},
  year={2023}
}

@article{PhysioNet-archehr-qa-bionlp-task-2025-1.3,
  author = {Soni, Sarvesh and Demner-Fushman, Dina},
  title = {{ArchEHR-QA: A Dataset for Addressing Patient's Information Needs related to Clinical Course of Hospitalization}},
  journal = {{PhysioNet}},
  year = {2026},
  month = jan,
  note = {Version 1.3},
  doi = {10.13026/n708-sn25},
  url = {https://doi.org/10.13026/n708-sn25}
}

@inproceedings{chen2024benchmarking,
  author = {Chen, Jiawei and Lin, Hongyu and Han, Xianpei and Sun, Le},
title = {Benchmarking large language models in retrieval-augmented generation},
year = {2024},
isbn = {978-1-57735-887-9},
publisher = {AAAI Press},
url = {https://doi.org/10.1609/aaai.v38i16.29728},
doi = {10.1609/aaai.v38i16.29728},
booktitle = {Proceedings of the Thirty-Eighth AAAI Conference on Artificial Intelligence and Thirty-Sixth Conference on Innovative Applications of Artificial Intelligence and Fourteenth Symposium on Educational Advances in Artificial Intelligence},
articleno = {1980},
numpages = {9},
series = {AAAI'24/IAAI'24/EAAI'24}
}

@article{zhang2025hallucination,
  title={Hallucination Mitigation for Retrieval-Augmented Large Language Models: A Review}, volume={13}, ISSN={2227-7390}, url={http://dx.doi.org/10.3390/math13050856}, DOI={10.3390/math13050856}, number={5}, journal={Mathematics}, publisher={MDPI AG}, author={Zhang, Wan and Zhang, Jing}, year={2025}, month=mar, pages={856} 
}

@misc{octofhir_fhirpath_rs,
  author       = {OctoFHIR},
  title        = {fhirpath-rs: FHIRPath tools written in Rust},
  howpublished = {\url{https://github.com/octofhir/fhirpath-rs}},
  note         = {Accessed: 2025-07-22},
}

\clearpage
\appendix
\renewcommand{\thefigure}{\thesection\arabic{figure}}
\renewcommand{\thetable}{\thesection\arabic{table}}
\setcounter{figure}{0}
\setcounter{table}{0}

\section{Additional Tables}
\label{app:tables}

\subsection{Comprehensive Performance Results}
\label{app:tab-combined}

See \autoref{tab:combined_performance}.

\begin{table*}[h]
\centering
\begin{tabular}{lc @{\extracolsep{12pt}} cc cc cc}
\hline
\textbf{Model} & \textbf{Base/FT} & \multicolumn{2}{c}{\textbf{Accuracy}} & \multicolumn{2}{c}{\textbf{Failure Rate}} & \multicolumn{2}{c}{\textbf{Acc. Excl. Failures}} \\ [0ex] \cline{3-8}
\rule{0pt}{1ex} & & \textit{Clinical} & \textit{Patient} & \textit{Clinical} & \textit{Patient} & \textit{Clinical} & \textit{Patient} \\ [0ex] 
\rule{0pt}{0ex}\textit{a) Retrieval} & & &  & & & \\ \hline
4.1-mini &Base & 0.42 & 0.41 & 0.17 & 0.14 & 0.50 & 0.48 \\
4o-mini &Base  & 0.24 & 0.15 & 0.46 & 0.40 & 0.45 & 0.25 \\
o4-mini &Base  & 0.33 & 0.33 & 0.36 & 0.34 & 0.51 & 0.49 \\
\rule{0pt}{0ex}\textit{b) Query-first} & & &  & & & \\ \hline
4.1-mini &Base & 0.25 & 0.27 & 0.15 & 0.23 & 0.29 & 0.35 \\
4.1-mini &FT-10   & 0.68 & 0.61 & 0.02 & 0.10 & 0.69 & 0.67 \\
4.1-mini &FT-30   & 0.72 & 0.64 & 0.01 & 0.09 & 0.73 & 0.70 \\
4.1-mini &FT-100  & 0.73 & 0.66 & 0.01 & 0.09 & 0.74 & 0.73 \\
4o-mini &Base  & 0.25 & 0.28 & 0.02 & 0.09 & 0.26 & 0.31 \\ 
4o-mini &FT-10    & 0.70 & 0.62 & 0.01 & 0.09 & 0.70 & 0.68 \\
4o-mini &FT-30    & 0.70 & 0.63 & 0.02 & 0.09 & 0.71 & 0.69 \\
4o-mini &FT-100   & 0.73 & 0.66 & 0.01 & 0.09 & 0.74 & 0.73 \\
4.1-nano &Base & 0.02 & 0.07 & 0.22 & 0.29 & 0.02 & 0.09 \\
4.1-nano &FT-10   & 0.59 & 0.49 & 0.04 & 0.11 & 0.61 & 0.55 \\
4.1-nano &FT-30   & 0.68 & 0.54 & 0.04 & 0.11 & 0.71 & 0.61 \\
4.1-nano &FT-100  & 0.67 & 0.57 & 0.01 & 0.11 & 0.68 & 0.63 \\
o4-mini &Base  & 0.35 & 0.38 & 0.09 & 0.15 & 0.39 & 0.45 \\ 
\hline
\end{tabular}
\caption{Comprehensive experimental results of retrieval and query-first strategies across clinical and patient perspectives for all models evaluated on the test split. FT-X indicates the fine-tuning dataset included X patients.}
\label{tab:combined_performance}
\end{table*}

\subsection{Comprehensive Usage Results}

See \autoref{tab:token_efficiency_all_models}.

\begin{table*}[t]
\centering
\begin{tabular}{lc @{\extracolsep{12pt}} cc}
\hline
\textbf{Model} & \textbf{Base/FT} & \multicolumn{2}{c}{\textbf{Token Usage}} \\ [0ex] \cline{3-4}
\rule{0pt}{1ex} & & \textit{Tokens / Question} & \textit{STD} \\ [0ex]
\rule{0pt}{0ex}\textit{a) Retrieval} & & & \\ \hline
4.1-mini & Base & 577,281 & 1,889,508 \\
4o-mini  & Base & 737,846 & 1,992,794 \\
o4-mini  & Base & 629,839 & 1,926,723 \\
\rule{0pt}{0ex}\textit{b) Query-first} & & & \\ \hline
4.1-mini & Base   & 697   & 24 \\
4.1-mini & FT-10  & 704   & 25 \\
4.1-mini & FT-30  & 704   & 25 \\
4.1-mini & FT-100 & 703   & 25 \\
4o-mini  & Base   & 692   & 18 \\
4o-mini  & FT-10  & 703   & 25 \\
4o-mini  & FT-30  & 704   & 25 \\
4o-mini  & FT-100 & 703   & 25 \\
4.1-nano & Base   & 684   & 19 \\
4.1-nano & FT-10  & 703   & 25 \\
4.1-nano & FT-30  & 703   & 25 \\
4.1-nano & FT-100 & 702   & 23 \\
o4-mini  & Base   & 1,609 & 420 \\ \hline
\end{tabular}
\caption{Token consumption per question on the test split across retrieval and query-first strategies for all models evaluated. FT-X indicates the fine-tuning dataset included X patients.}
\label{tab:token_efficiency_all_models}
\end{table*}

\subsection{FHIRPath Query Match Rates}

See \autoref{tab:exact-query-match-accuracy}.

\begin{table*}[t]
\centering
\begin{tabular}{lcc}
\hline
\textbf{Model (Train Size)} & \textbf{Query Match Rate}& \textbf{Query Match Rate with Masked Numeric Quantities}\\ \hline
4.1-mini (Base) & 0.01 & 0.01 \\
4.1-mini (10)   & 0.31 & 0.58 \\
4.1-mini (30)   & 0.35 & 0.63 \\
4.1-mini (100)  & 0.34 & 0.63 \\ 
4.1-nano (Base) & 0.00 & 0.01 \\
4.1-nano (10)   & 0.07 & 0.39 \\
4.1-nano (30)   & 0.14 & 0.44 \\
4.1-nano (100)  & 0.16 & 0.43 \\
4o-mini (Base)  & 0.03 & 0.05 \\
4o-mini (10)    & 0.24 & 0.58 \\
4o-mini (30)    & 0.33 & 0.63 \\
4o-mini (100)   & 0.35 & 0.64 \\ 
o4-mini (Base)  & 0.03 & 0.03 \\
\hline
\end{tabular}
\caption{The query match rate with masked numeric quantities consistently increased after fine-tuning, demonstrating that a high number of query inaccuracies were due to incorrect date filtering in otherwise correct queries. This metric was computed by masking all numeric substrings (i.e. dates, patient ids, dose quantities) in the generated and ground truth query before comparing strings.}
\label{tab:exact-query-match-accuracy}
\end{table*}

\section{Additional Figures}
\label{app:figures}

\subsection{Token Usage by Patient EHR Size}
\label{app:fig-efficiency}

\begin{figure}[h]
    \centering
    \includegraphics[width=0.95\linewidth]{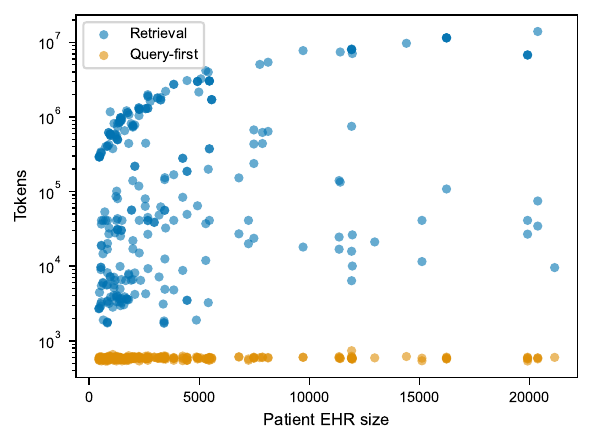}
    \caption{Patient record size vs token usage (log scale) using (o4-mini). The upper bound on tokens used in the retrieval-first approach scales linearly with the patient EHR size, whereas query-first remains constant. Retrieval tokens corresponds to total tokens or attempted tokens if the context length limit was exceeded.}
    \label{fig:efficiency}
\end{figure}

See \autoref{fig:efficiency}.

\subsection{Paraphrase Statistics}
\label{app:paraphrase-stats}

\begin{figure*}[t]
    \centering
    \includegraphics[width=0.95\linewidth]{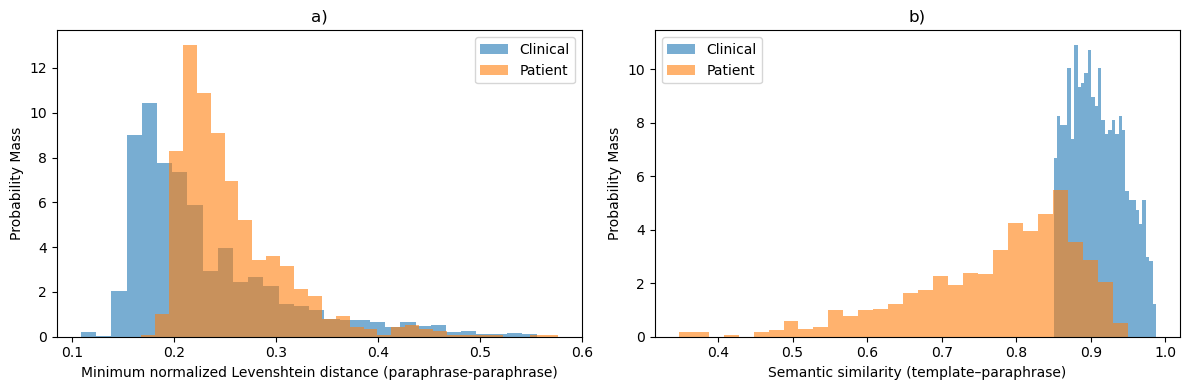}
    \caption{Histograms of a) the minimum normalized Levenshtein distance between paraphrase and each other paraphrase of the same question template and b) the semantic similarity of paraphrase to original question template by perspective.}
    \label{fig:paraphrase-stats}
\end{figure*}

See \autoref{fig:paraphrase-stats}.

\section{Error Analysis Examples}
\label{app:examples}

\subsection{Example 1}
\label{ssub:example_1}
\textbf{Model:} query\_gen-4.1-nano\_100 \\
\textbf{Question:} until last year, for patient 10003400, what was the last time a medication was prescribed using the IV route? \\
\textbf{Context:} Assume current datetime is 2137-09-02T19:17:11 \\
\textbf{true\_query:} {\footnotesize \seqsplit{Bundle.entry.resource.where(resourceType='MedicationRequest' and dosageInstruction.route.coding.code = 'IV' and authoredOn.toDateTime() <= @2136-12-31T19:17:11.000Z).sort(authoredOn).last().authoredOn.toDateTime()}} \\
\textbf{model\_query:} {\footnotesize \seqsplit{Bundle.entry.resource.where(resourceType='MedicationRequest' and dosageInstruction.route.coding.code = 'IV' and authoredOn.toDateTime() <= @2137-12-31T19:17:11.000Z).sort(authoredOn).last().authoredOn.toDateTime()}} \\
\textbf{true\_answer:} 2136-12-14T08:52:22-05:00 \\
\textbf{model\_answer:} 2137-09-02T13:45:41-04:00 \\
\textbf{Analysis:} The model used the incorrect date in the generated query despite having an exact match otherwise.
\vspace{1em}

\subsection{Example 2}
\label{ssub:example_2}
\textbf{Model:} query\_gen-4o-mini\_100 \\
\textbf{Question:} Can you tell me what the last careunit was until 2144? \\
\textbf{Context:} Assume current datetime is 2145-01-03T20:33:00 \\
\textbf{true\_query:}{ \footnotesize \seqsplit{Bundle.entry.resource.where(resourceType='Encounter' and select((period.start | period.end).sort().first().toDateTime()) <= @2144-12-31T05:02:42.000Z).sort(period.start).last().location.sort(period.start).last().location.resolve().name}} \\
\textbf{model\_query:}{\footnotesize \seqsplit{Bundle.entry.resource.where(resourceType='Location' and name.exists() and period.start.toDateTime() <= @2144-12-31T23:25:29.000Z).sort(period.start).last().name}} \\
\textbf{true\_answer:} Cardiac Vascular Intensive Care Unit (CVICU) \\
\textbf{model\_answer:} [] \\
\textbf{Analysis:} The generated query does not demonstrate knowledge about the structure of a Location or Encounter resource. Location resources contain no temporal information; they are linked to Encounters which contain visit information including the time period.
\vspace{1em}

\subsection{Example 3}
\label{ssub:example_3}
\textbf{Model:} query\_gen-4.1-mini\_100 \\
\textbf{Question:} I'd like to know the number of times I took Potassium Chloride until 20/05/2187. \\
\textbf{Context:} Assume current datetime is 2187-05-23T17:20:00 \\
\textbf{true\_query:} {\footnotesize \seqsplit{Bundle.entry.resource.where(resourceType='MedicationAdministration' and context.resolve().identifier.system='http://fhir.mimic.mit.edu/identifier/encounter-icu' and medicationCodeableConcept.coding.display = 'Potassium Chloride' and select((effectiveDateTime | effectivePeriod.start).sort().first().toDateTime()) <= @2187-05-20T18:21:11.000Z).count()}} \\
\textbf{model\_query:} {\footnotesize \seqsplit{Bundle.entry.resource.where(resourceType='MedicationRequest' and (medicationReference.resolve().identifier.where(value='Potassium Chloride').exists() or medicationCodeableConcept.coding.code='Potassium Chloride') and authoredOn.toDateTime() <= @2187-05-20T19:10:10.000Z).count()}} \\
\textbf{true\_answer:} 1 \\
\textbf{model\_answer:} 0 \\
\textbf{Analysis:} The patient-paraphrased question contains ambiguity in the phrase "I took Potassium Chloride". The incorrectly assumes Potassium Chloride was a prescription, but it was intended to be a question about drug intake while an in-patient. The clinician-paraphrased question would clearly specify if the drug was 'prescribed' or 'administered'.
\vspace{1em}

\subsection{Example 4}
\label{ssub:example_4}
\textbf{Model:} query\_gen-4.1-mini\_100 \\
\textbf{Question:} I need to know how many medications were issued to me until 4 months ago. \\
\textbf{Context:} Assume current datetime is 2151-01-15T23:17:00 \\
\textbf{true\_query:} {\footnotesize \seqsplit{Bundle.entry.resource.where(resourceType='MedicationRequest' and authoredOn.toDateTime() <= @2150-09-30T23:17:00.000Z).select(medicationReference.resolve() | medicationCodeableConcept).distinct().count()}} \\
\textbf{model\_query:} {\footnotesize \seqsplit{Bundle.entry.resource.where(resourceType='MedicationRequest' and authoredOn.toDateTime() <= @2150-09-30T23:17:00.000Z).count()}} \\
\textbf{true\_answer:} 258 \\
\textbf{model\_answer:} 1442 \\
\textbf{Analysis:} The generated query counts the total number of prescriptions instead of counting distinct medication types as the question required. The result is a partially correct query.
\vspace{1em}

\subsection{Example 5}
\label{ssub:example_5}
\textbf{Model:} query\_gen-4o-mini\_base \\
\textbf{Question:} Which microbiology test did patient 10003046 undergo last since 14 months ago? \\
\textbf{Context:} Assume current datetime is 2154-01-09T11:53:00 \\
\textbf{true\_query:} {\footnotesize \seqsplit{Bundle.entry.resource.where(resourceType='Observation' and code.coding.system = 'http://fhir.mimic.mit.edu/CodeSystem/microbiology-test' and effectiveDateTime.toDateTime() >= @2152-11-01T11:53:00.000Z).sort(effectiveDateTime).last().code.coding.display}} \\
\textbf{model\_query:} {\footnotesize \seqsplit{Bundle.entry.resource.where(resourceType='Observation' and code.coding.display = 'Microbiology' and effectiveDateTime.toDateTime() >= @2152-11-01T11:53:00.000Z).sort(effectiveDateTime).last().code.coding.display}} \\
\textbf{true\_answer:} C. difficile PCR \\
\textbf{model\_answer:} "" \\
\textbf{Analysis:} The generated query fails to specify a microbiology test code system. It incorrectly applies a generic display filter that does not exist, missing the required microbiology result. Apart from this filter clause, the query is correct.
\vspace{1em}

\subsection{Example 6}
\label{ssub:example_6}
\textbf{Model:} query\_gen-4.1-mini\_base \\
\textbf{Question:} Provide the number of Condom Cath outputs associated with patient 10035631 since 10 years ago \\
\textbf{Context:} Assume current datetime is 2116-03-12T07:45:00 \\
\textbf{true\_query:} {\footnotesize \seqsplit{Bundle.entry.resource.where(resourceType='Observation' and code.coding.display = 'Condom Cath' and effectiveDateTime.toDateTime() >= @2106-01-01T07:45:00.000Z).count()}} \\
\textbf{model\_query:} {\footnotesize \seqsplit{Bundle.entry.resource.where(resourceType='Observation' and code.coding.display = 'Condom Cath' and effectiveDateTime.toDateTime() >= @2106-03-12T07:45:00.000Z).count()}} \\
\textbf{true\_answer:} 1 \\
\textbf{model\_answer:} 1 \\
\textbf{Analysis:} The generated query uses the incorrect date component, making the correct evaluated answer coincidental. 
\vspace{1em}

\subsection{Example 7}
\label{ssub:example_7}
\textbf{Model:} query\_gen-o4-mini\_base \\
\textbf{Question:} How many times was patient 10000032 given Sodium Polystyrene Sulfonate until 2180? \\
\textbf{Context:} Assume current datetime is 2180-08-10T22:48:00 \\
\textbf{true\_query:} {\footnotesize \seqsplit{Bundle.entry.resource.where(resourceType='MedicationRequest' and (medicationReference.resolve().identifier.where(value='Sodium Polystyrene Sulfonate').exists() or medicationCodeableConcept.coding.code='Sodium Polystyrene Sulfonate') and authoredOn.toDateTime() <= @2180-12-31T22:18:53.000Z).count()}} \\
\textbf{model\_query:} {\footnotesize \seqsplit{Bundle.entry.resource.where(resourceType='MedicationAdministration' and medicationCodeableConcept.coding.display = 'Sodium Polystyrene Sulfonate' and effectiveDateTime.toDateTime() <= @2180-12-31T22:48:00.000Z).count()}} \\
\textbf{true\_answer:} 4 \\
\textbf{model\_answer:} 0 \\
\textbf{Analysis:} The generated query has no correct components. Likely stemming from ambiguity in the question with the phrase "given Sodium Polystyrene Sulfonate", the model assumes this to be a drug administered in a hospital context.
\vspace{1em}

\end{document}